# Pelacakan Intensitas Piksel untuk Pemantauan Pernapasan Jarak Jauh: Sebuah Studi pada Subjek Orang Indonesia

*Pixel Intensity Tracking for Remote Respiratory Monitoring: A Study on Indonesian Subject*

Muhammad Yahya Ayyashy Mujahidan [1)], Martin Clinton Tosima Manullang [*,1)]

[1)]*Program Studi Teknik Informatika, Fakultas Teknologi Informasi, Institut Teknologi Sumatera*
*Lampung Selatan, Lampung, Indonesia*

*Abstract - Respiratory rate is a vital sign indicating various health conditions. Traditional contact-based measurement methods are often uncomfortable, and alternatives like respiratory belts and smartwatches have limitations in cost and operability. Therefore, a non-contact method based on Pixel Intensity Changes (PIC) with RGB camera images is proposed. Experiments involved 3 sizes of bounding boxes, 3 filter options (Laplacian, Sobel, and no filter), and 2 corner detection algorithms (ShiTomasi and Harris), with tracking using the Lukas-Kanade algorithm. Eighteen configurations were tested on 67 subjects in static and dynamic conditions. The best results in static conditions were achieved with the Medium Bounding box, Sobel Filter, and Harris Method (MAE: 0.85, RMSE: 1.49). In dynamic conditions, the Large Bounding box with no filter and ShiTomasi, and Medium Bounding box with no filter and Harris, produced the lowest MAE (0.81) and RMSE (1.35).*

**Keywords -** *respiratory rate; health monitoring; pixel intensity changes; rgb image; motion detection*

*Abstrak – Laju pernapasan adalah tanda vital yang menunjukkan berbagai kondisi kesehatan. Metode pengukuran tradisional yang melibatkan kontak fisik sering kali tidak nyaman, dan alternatif seperti respiratory belt dan smartwatch memiliki keterbatasan biaya dan operabilitas. Oleh karena itu, diusulkan metode non-kontak berbasis Perubahan Intensitas Piksel (PIC) dengan gambar dari kamera RGB. Percobaan melibatkan 3 ukuran bounding box, 3 opsi filter (Laplacian, Sobel, dan tanpa filter), serta 2 algoritma deteksi sudut (ShiTomasi dan Harris), dengan pelacakan menggunakan algoritma Lukas-Kanade. Delapan belas konfigurasi diuji pada 67 subjek dalam kondisi statis dan dinamis. Hasil terbaik dalam kondisi statis dicapai dengan kombinasi Medium Bounding box, Filter Sobel, dan Metode Harris (MAE: 0,85, RMSE: 1,49). Dalam kondisi dinamis, kombinasi Large Bounding box tanpa filter dan ShiTomasi, serta Medium Bounding box tanpa filter dan Harris, menghasilkan MAE terendah (0,81) dan RMSE terendah (1,35).*

**Kata Kunci -** *laju pernapasan; pemantauan kesehatan; perubahan intensitas piksel; citra rgb; deteksi pergerakan*

---
[*)] Corresponding author (Martin C.T. Manullang)
Email: martin.manullang@if.itera.ac.id
*Both authors contributed equally to this work*

## I. INTRODUCTION

Respiratory rate (RR) is a vital sign indicative of various health conditions. Abnormal RR can signal the presence of several diseases, as research has shown that RR is highly responsive to numerous health conditions, including abnormal cardiac activity, pneumonia, and various stressors such as emotional stress, cognitive load, environmental temperature, and physical activity. Beyond its medical relevance, RR is also a crucial parameter in assessing sports performance, particularly in fitness and cardiovascular training. Monitoring RR during exercise provides valuable insights into an individual's endurance, aerobic capacity, and overall physical conditioning. The extensive range of conditions influenced by RR underscores its critical role in both health monitoring and sports performance evaluation, highlighting its importance in maintaining overall human well-being [1].

Traditionally, respiration is measured through direct contact methods. These methods typically involve analyzing parameters derived from the airflow during respiration using devices like spirometers and chest and abdominal motion sensors. Such methods necessitate physical contact with the patient, potentially causing skin irritation and discomfort in short and long-term use [2]. Additionally, alternative devices like respiratory belts and smartwatches exist for RR monitoring. Yet, they also pose discomfort during prolonged use and are often inaccessible to lower socioeconomic groups due to their cost [3].

Non-contact monitoring systems are considered a solution to the discomfort and inefficiency associated with contact-based vital sign monitoring systems. Various non-contact alternatives have been explored, including Doppler Radar [4], commonly used in automotive applications, three-dimensional time-of-flight sensors capturing subject movement in 3D [5], and Impulse-Radio Ultra-Wideband (IR-UWB), which emits short-range radio frequency pulses across a wide frequency band [2]. Despite the promising performance of these alternatives, challenges remain, such as high costs and the need for skilled operators [5].

A proposed solution, discussed in the research conducted by Massaroni et al. [6], involves using multiple resolutions on a high-quality RGB camera equipped with a Charge-Coupled Device (CCD) image sensor to capture subject chest movement recording. The RGB information in the recording is then processed to form a signal representing the y-axis movement pattern of the subject chest wall. This method of leveraging body position changes is portable and practical, with a good MAE result on the HD resolution (720p) benchmark of 0.55. However, CCD-equipped cameras are mostly expensive and inaccessible among the masses.

Gwak et al. [7] developed a method for monitoring respiratory rate (RR) and detecting breathing absence by tracking head movements in RGB video, as respiratory motion affects the head position. Using the Kanade-Lucas-Tomasi (KLT) tracker, this approach achieved a Mean Absolute Error (MAE) of 1.91 to 2.46 bpm and a Root Mean Squared Error (RMSE) of 2.7 to 3.54 bpm, though further accuracy improvements are possible. Similarly, Hassan et al. [8] proposed using Photoplethysmography (PPG) with a periodicity-based voting scheme (PVS), focusing on specific facial regions. While practical in uncontrolled environments, this method had an RMSE of 5.2 bpm, indicating room for enhancement.

Notably, most of the experiments in noncontact respiratory monitoring have not involved Indonesian datasets, limiting the generalizability of these methods to populations with different physical and environmental characteristics. Commonly used datasets for noncontact respiratory monitoring typically include the MIT-BIH Polysomnographic Database done by Fan et al. [9], Capnobase done by Karlen et al. [10], and the Apnea-ECG database done by Sadr et al. [11]. These datasets predominantly consist of Western or mixed-population subjects, leaving a gap in research on how these techniques perform across diverse ethnic groups, including Indonesians. Expanding the scope of datasets to include varied populations could help refine these methods and improve their accuracy in different demographic contexts

To address these issues, we propose a low-cost, non-contact RR measurement method using Pixel Intensity Tracking (PIT) from RGB camera video. This study uniquely involves the collection of our own dataset of half-body RGB images, specifically focusing on the region from the head to the lower chest, sourced from an Indonesian population. By detecting movements and analyzing the influence of various features on RR readings, our approach aims to achieve highly accurate results with minimal error comparable to clinical measurements. The use of a locally sourced dataset adds a novel dimension to this research, enhancing its relevance for diverse populations and advancing the accuracy of non-contact RR monitoring methods.

## II. METHODOLOGY

To enhance the performance of the proposed non-contact respiratory rate (RR) measurement method, several preprocessing techniques can be applied to improve the quality of the Region of Interest (ROI) in the RGB images. Since the accuracy of the method depends heavily on movement detection within the ROI, reducing noise and enhancing relevant motion points are critical steps [8]. One effective approach is the application of the Laplacian filter, which acts as a diffusion operator by spreading intensity values across pixels, resulting in noise reduction and image smoothing, particularly around edges [12]. This filter is widely used in image processing to emphasize corners and edges by attenuating noise, as it maps rapid intensity changes that characterize such features. The corners detected through this method are then sharpened, leading to an increase in intensity at pixels identified as corners [13].

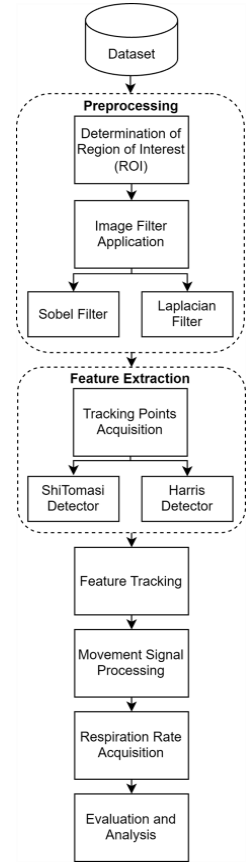

**Figure 1.** Research Stages

Alternatively, the Sobel filter offers another technique to improve ROI quality. The Sobel operator applies two convolution kernels to the image, effectively highlighting edges and enhancing the detection of key features within the ROI [14]. Once preprocessing is complete, corner detection methods can be employed to identify tracking points within the ROI. For this, the algorithm by Shi and Tomasi [15] is a viable option, as it uses a spatial gradient to detect corners based on the uniformity of neighboring pixel values. It identifies a corner when two tangent edges form a consistent gradient. To compare performance, the Harris and Stephens algorithm [16] offers another approach, where each pixel and its neighbors are analyzed, and corners are detected when two lines of pixels form an edge.

The combination of these preprocessing and corner detection techniques, illustrated in Figure 1, provides an efficient and accessible solution for optimizing motion detection in RR monitoring, particularly suited for low-resource environments. More details about the processing pipeline and this research methodology will be discussed on the following subsection.

### A. Dataset

The dataset used in this study comprises pre-processed data, including both respiratory rates and video recordings of subjects captured under varying conditions. It includes 67 participants (male and female), aged between 20 and 25 years, representing a demographic of physically healthy young adults. The subjects were recorded in two distinct conditions: a static condition, where heart rates were maintained between 60-100 bpm, and a dynamic condition. In the dynamic condition, subjects were asked to perform physical exercise until their heart rate exceeded 120 bpm, followed by continuous data recording. This approach ensured that vital signs, such as respiratory rate, were captured during a period of recovery, allowing for the monitoring of dynamic physiological changes as the subjects' heart rates gradually returned to baseline.

The dataset was collected by measuring the respiratory rate of the subjects using a respiratory belt placed around the chest while they were seated. This method provided ground truth (GT) data for later use in evaluation metrics. The respiratory belt had a sampling rate of 20 Hz. Simultaneously, video recordings of the subjects were captured while their respiratory rates were being measured, with a resolution of 720p (1280x720) and a frame rate of 30 frames per second (fps). The data collection process for each subject lasted 1 minute.

### B. ROI Localization

In the processing pipeline, the first step is to define the Region of Interest (ROI) as the focal area of the image to be processed for respiratory rate extraction. The bounding box for the chest region was defined by applying heuristic scaling and positional adjustments to the coordinates of the face detection bounding box obtained from Dlib's frontal face detector. For the benchmark purpose, we set three different sizes of bounding box. The illustration can be found in Figure 2.

After obtaining the bounding box, we applied filters to the image within the defined area. In this phase of the study, three options were explored: the Laplacian filter, the Sobel filter, and a control option with no filter applied. These options were used to evaluate the impact of filter application on the accuracy of respiratory rate detection. Figure 3 provides a visualization of the applied filters using the medium ROI configuration. The aim of implementing these filters is to enhance the feature inside the ROI.

### C. Laplacian Filter

The first filter option is the Laplacian filter. It works as a second-order derivative operator that uses a single convolution kernel to detect regions of rapid intensity change (edges) in an image. The operator computes the second derivative of the image intensity function, capturing both horizontal and vertical changes simultaneously and one 3x3 kernel, such as in (1) and (2) [17]. The advantage of the Laplacian method is its ability to highlight fine details in the image, making it sensitive to both bright-to-dark and dark-to-bright transitions [13].

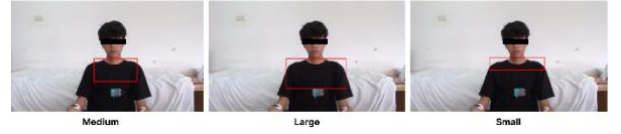

**Figure 2.** Size of ROIs Used

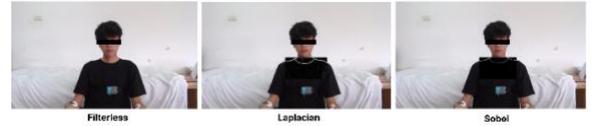

**Figure 3.** Filters Application Visualisation

$$\nabla^2 f(x,y) = \frac{\partial^2 f}{\partial x^2} + \frac{\partial^2 f}{\partial y^2} \quad (1)$$

$$G_{xy} = \begin{bmatrix} 0 & 1 & 0 \\ 1 & -4 & 1 \\ 0 & 1 & 0 \end{bmatrix} \quad (2)$$

### D. Sobel Filter

Another filter option we assessed is the Sobel operator, which computes image gradients in both the horizontal and vertical directions. These gradients are obtained by processing the pixel values using two distinct 3x3 convolution kernels, as defined in equations (3) and (4) [18]. The strength of the Sobel operator lies in its ability to perform edge detection while simultaneously reducing noise, as the kernel inherently smooths the image before highlighting edges [14].

$$G = \sqrt{G_x^2 + G_y^2} \quad (3)$$

$$G_x = \begin{bmatrix} -1 & 0 & 1 \\ -2 & 0 & 2 \\ -1 & 0 & 1 \end{bmatrix} G_y = \begin{bmatrix} 1 & 2 & 1 \\ 0 & 0 & 0 \\ -1 & -2 & -1 \end{bmatrix} \quad (4)$$

Given the various combinations of bounding boxes and filters to be evaluated, we created acronyms for each experimental setup to simplify references throughout the study. Table 1 lists all the options along with their corresponding acronyms.

**Table 1.** Acronyms for all configuration options used in the research

| Filter | Bbox Size | Method Name | Acronym |
|---|---|---|---|
| Filterless | Bbox Medium | Filterless-Bbox Medium | FLBM |
| Filterless | Bbox Large | Filterless-Bbox Large | FLBL |
| Filterless | Bbox Small | Filterless-Bbox Small | FLBS |
| Laplacian | Bbox Medium | Laplacian-Bbox Medium | LPBM |
| Laplacian | Bbox Large | Laplacian-Bbox Large | LPBL |
| Laplacian | Bbox Small | Laplacian-Bbox Small | LPBS |
| Sobel | Bbox Medium | Sobel-Bbox Medium | SOBM |
| Sobel | Bbox Large | Sobel-Bbox Large | SOBL |
| Sobel | Bbox Small | Sobel-Bbox Small | SOBS |

### E. Feature Extraction

After applying the filter, the next step is to define the feature points within the ROI. Each configuration will be processed using two corner detection methods: the Harris detector and the Shi-Tomasi detector.

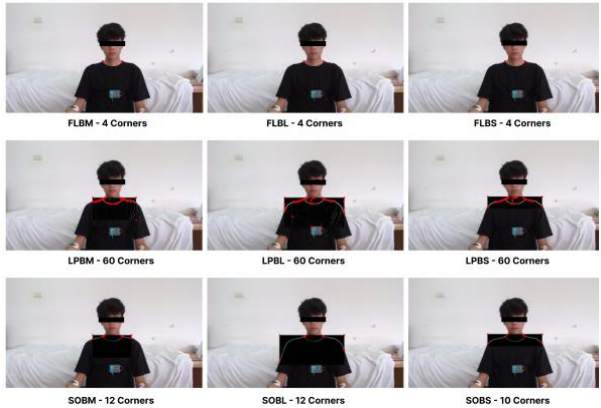

**Figure 4.** Tracking points visualisation for ShiTomasi Detector

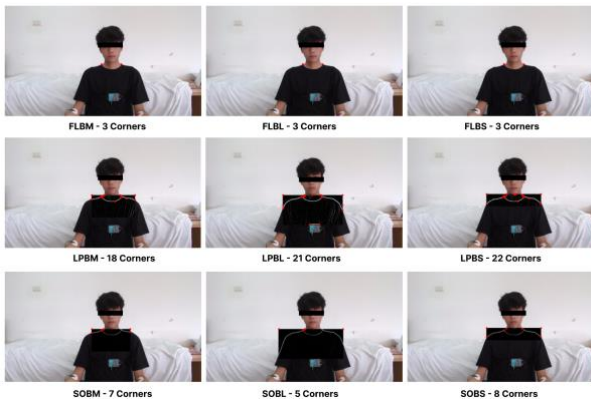

**Figure 5.** Tracking points visualisation for Harris Detector

These methods enable a thorough analysis of key points, contributing to improved accuracy in motion detection in the later stages of the study. The resulting tracking points for each configuration are shown in Figure 4 for the Shi-Tomasi detector and Figure 5 for the Harris detector

### F. Harris Corner Detector

The Harris corner detector works by using the autocorrelation matrix (M) to capture intensity changes in both the x and y directions. These changes are then processed through the corner response function, as described in equations (5) and (6) [16]. The strength of the Harris method lies in its ability to accurately identify corners while distinguishing them from edges and flat regions by analyzing local gradients. This makes the Harris detector a robust choice for corner detection in a wide range of image processing applications [19].

$$M = \sum \begin{bmatrix} I_x^2 & I_x I_y \\ I_x I_y & I_y^2 \end{bmatrix} \quad (5)$$

$$R = \det(M) - k \cdot (\text{trace}(M))^2 \quad (6)$$

### G. ShiTomasi Corner Detector

The next corner detection method, the ShiTomasi corner detector, works by analyzing the spatial gradient of pixel intensities. This method assesses the eigenvalues of the gradient matrix at each pixel (M), where corners are detected based on the minimum eigenvalue threshold as defined in (5) and (6) [15]. The advantage of the Shi-Tomasi detector is its robustness in distinguishing corners from edges, as it effectively measures the intensity variation in multiple directions, ensuring accurate corner localization even in noisy environments [20].

$$M = \sum \begin{bmatrix} I_x^2 & I_x I_y \\ I_x I_y & I_y^2 \end{bmatrix} \quad (7)$$

$$R = \min(\lambda_1, \lambda_2) \quad (8)$$

### H. Feature Tracking

After extracting the features within the ROI, these highlighted points are tracked using optical flow to generate a y-axis movement data signal. The Lucas-Kanade optical flow method is employed, which estimates the velocity of object motion between consecutive frames by using spatial and temporal image gradients. This method assumes that pixel intensities remain constant over time and computes the optical flow by solving a set of linear equations derived from the optical flow constraint equation, as defined in equation (9) [21]. The advantage of the Lucas-Kanade method lies in its ability to use a small patch around each pixel, applying a least-squares approach to solve for optical flow velocities. This reduces sensitivity to noise and enhances the robustness of motion estimation in local regions. By minimizing error within the local neighborhood, this method provides a more stable and reliable flow estimation [22].

$$I_x u + I_y v + I_t = 0 \quad (9)$$

### I. Movement Signal Processing

After acquiring the movement signal from the subjects in the previous step, the next stage involves filtering out irrelevant frequencies to reduce noise. First irrelevant corners are removed by filtering out the highest and lowest movement variance by 33%, then the acquired movement are filtered using a bandpass filter, with a low cutoff frequency of 0.1 and a high cutoff of 0.45. Figure 6 illustrates the visual differences between the unfiltered and filtered signals.

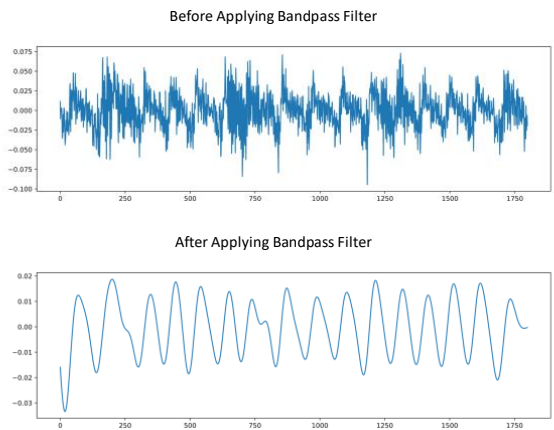

**Figure 6**. Visualisation of Movement Signal Filtering

A bandpass filter allows frequencies within a specific range to pass while attenuating frequencies outside that range. The advantage of using a bandpass filter is its ability to isolate and enhance the signals of interest by filtering out both low-frequency noise and high-frequency interference [23]. Following filtering, the next step is to normalize both the movement and ground truth signals using z-normalization to improve signal clarity, as shown in Figure 7.

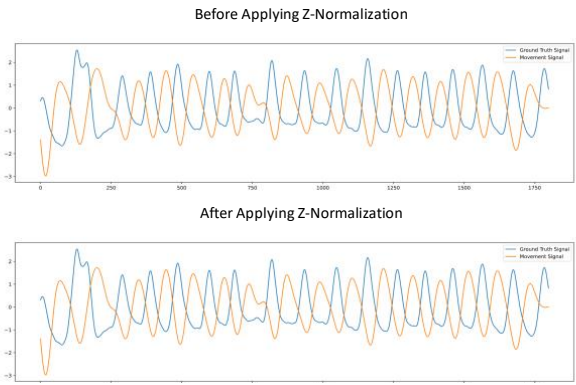

**Figure 7.** Visualisation of Z-Normalization Application in Movement Signal

Z-normalization standardizes the data by subtracting the mean and dividing by the standard deviation, as defined in equation (10). The benefit of this method is that it transforms data to have a mean of 0 and a standard deviation of 1, making it easier to compare features with different scales and enhancing the overall readability of the data [24].

$$Z\text{-Score} = \frac{Length\ of\ the\ data\ row\ -\ mean\ of\ the\ data}{standard\ deviation\ of\ the\ data} \quad (10)$$

### J. Respiration Rate Acquisition

With signal processing complete, the next step is to analyze the processed signals. Respiration rates were determined by identifying the peaks in the signal, with local maxima used to detect these peaks. A visualization of the detected peaks is shown in Figure 8.

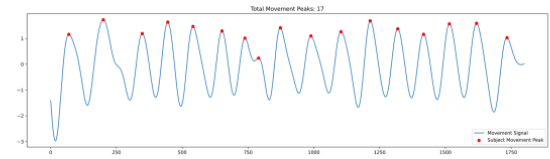

**Figure 8.** Visualisation of Peaks Detected in The Movement Signal

A local maximum is a point in the dataset where the value is higher than its neighboring points, identified using the first and second derivatives, as described in equation (11). The advantage of detecting local maxima lies in its ability to pinpoint significant peaks in the processed data, providing a reliable method for identifying key events such as respiration cycles [25].

$$f'(c) = first\ derivative$$
$$f''(c) = second\ derivative$$

$$if\ f'(c) = 0\ and\ f''(c) < 0,\ then\ f(x)\ has\ a\ local\ maximum\ at\ x = c. \quad (11)$$

## III. RESULTS AND DISCUSSION

In order to evaluate this research, Mean Average Error (MAE), Root Mean Squared Error (RMSE) and standard deviation were selected as the performance metrics of acquired respiration rates from every combinations of method compared to ground truth data taken with the respiratory belt device, with the subject both static and dynamic respiration state.

The first performance metric, MAE, was used to identify the most accurate method, with the lowest MAE indicating the best result. For static respiration states, as shown in Table 2, the combination of a Sobel filter, medium-sized bounding box, and Harris corner detector achieved the lowest MAE among all tested configurations. This indicates that the choice of components significantly impacts performance.

Regarding bounding box size, the medium-sized option consistently produced the best results in static conditions, with an average MAE of 1.01, outperforming both large (1.04) and small-sized (1.14) bounding boxes. In terms of filtering techniques, the Sobel filter was the most effective, yielding an average MAE of 0.99, compared to 1.14 for the Laplacian filter and 1.03 for the filterless option.

**Table 2.** Performance Results of Proposed Combinations of Methods on Static Dataset

| METHOD | PR | | |
|---|---|---|---|
| | MAE | RMSE | SD |
| ShiTomasi - FLBM | 1.01 | 1.72 | 1.71 |
| ShiTomasi - FLBL | 0.96 | 1.7 | 1.7 |
| ShiTomasi - FLBS | 1.09 | 2.05 | 2.05 |
| ShiTomasi - LPBM | 1.12 | 2.18 | 2.18 |
| ShiTomasi - LPBL | 1.09 | 2.22 | 2.22 |
| ShiTomasi - LPBS | 1.21 | 2.22 | 2.22 |
| ShiTomasi - SOBM | 0.96 | 1.61 | 1.59 |
| ShiTomasi - SOBL | 0.94 | 1.63 | 1.61 |
| ShiTomasi - SOBS | 0.96 | 1.6 | 1.6 |
| Harris - FLBM | 1.01 | 1.58 | 1.57 |
| Harris - FLBL | 0.96 | 1.62 | 1.61 |
| Harris - FLBS | 1.09 | 2.12 | 2.12 |
| Harris - LPBM | 1.12 | 2.15 | 2.15 |
| Harris - LPBL | 1.09 | 2.08 | 2.05 |
| Harris - LPBS | 1.21 | 2.38 | 2.36 |
| Harris - SOBM | 0.96 | **1.49** | 1.48 |
| Harris - SOBL | **0.94** | 1.98 | 1.98 |
| Harris - SOBS | 0.96 | 1.98 | 1.98 |

**Table 3.** Performance Results of Proposed Combinations of Methods on Dynamic Dataset

| METHOD | PR | | |
|---|---|---|---|
| | MAE | RMSE | SD |
| ShiTomasi - FLBM | 0.88 | 1.45 | 1.42 |
| ShiTomasi - FLBL | 0.84 | **1.35** | 1.29 |
| ShiTomasi - FLBS | 0.9 | 1.36 | 1.33 |
| ShiTomasi - LPBM | 0.87 | 1.44 | 1.38 |
| ShiTomasi - LPBL | 0.93 | 1.53 | 1.62 |
| ShiTomasi - LPBS | 1.03 | 1.53 | 1.57 |
| ShiTomasi - SOBM | 0.9 | 1.4 | 1.37 |
| ShiTomasi - SOBL | 0.85 | 1.51 | 1.49 |
| ShiTomasi - SOBS | 0.88 | 1.48 | 1.44 |
| Harris - FLBM | **0.81** | 1.36 | 1.31 |
| Harris - FLBL | 0.84 | **1.35** | 1.29 |
| Harris - FLBS | 0.9 | 1.47 | 1.43 |
| Harris - LPBM | 0.94 | 1.55 | 1.49 |
| Harris - LPBL | 0.85 | 1.44 | 1.38 |
| Harris - LPBS | 0.96 | 1.56 | 1.56 |
| Harris - SOBM | 0.88 | 1.36 | 1.33 |
| Harris - SOBL | 0.87 | 1.38 | 1.35 |
| Harris - SOBS | 0.88 | 1.48 | 1.45 |

The optimal configuration—medium bounding box with Sobel filter—achieved an MAE of 0.85 when paired with the Shi-Tomasi corner detector. Meanwhile while comparing the effectiveness of the filters, the lowest average MAE were produced by the sobel filter (0.99) compared to the laplacian filter and filterless option with the average MAEs of 1.14 and 1.03 respectively.

or the dynamic respiration dataset (Table 3), the optimal combination shifted. The lowest MAE was achieved with a large-sized bounding box, filterless option, and Harris corner detector, yielding an MAE of 0.86. This demonstrates how dataset characteristics influence performance. In this case, the large bounding box outperformed the medium and small-sized ones, which had MAEs of 0.88 and 0.93, respectively.

Similarly, the filterless option proved more effective in dynamic conditions, with an average MAE of 0.86, compared to 0.87 for the Sobel filter and 0.93 for the Laplacian filter. However, the best overall result (MAE of 0.81) was obtained with the combination of a medium-sized bounding box, the filterless option, and the Harris corner detector, highlighting the importance of optimal component selection in different scenarios.

For the results of the RMSE evaluation metric shown on Table 2 and Table 3, highlights the combination of the medium-sized bounding box, Sobel filter, and Harris corner detector yielded the lowest RMSE of 1.49 for static respiration. On the dynamic dataset, the combination of a large bounding box, filterless option, and Harris corner detector achieved the lowest RMSE of 1.35.

The standard deviation (SD) results reveal distinct patterns in the variability of prediction performance across different method combinations. In the static dataset, the combination of the small-sized bounding box, Laplacian filter, and ShiTomasi corner detection algorithm exhibited the highest variability, indicating greater inconsistency in the predicted respiration rates. Conversely, the combination of the medium-sized bounding box, Sobel filter, and Harris corner detector proved to be more stable, with lower standard deviation, suggesting more reliable and consistent predictions.

For the dynamic dataset, a similar trend was observed. The small-sized bounding box, Laplacian filter, and ShiTomasi corner detector once again showed higher variability in prediction performance on average, reflecting less stability. On the other hand, the combination of the large-sized bounding box, Sobel filter, and Harris corner detector produced lower standard deviations, indicating a more stable and consistent prediction performance under dynamic conditions. These results suggest that certain method combinations consistently offer more reliable predictions by minimizing variability in both static and dynamic respiration states..

### IV. CONCLUSION

In conclusion, the evaluation using MAE, RMSE, and standard deviation metrics has demonstrated the significance of optimal component selection in predicting respiration rates accurately and consistently. For static respiration state dataset, the medium-sized bounding box with a Sobel filter and Harris corner detector yielded the best accuracy, with the lowest MAE (0.94) and the large-sized bounding box with a Sobel filter and Harris corner detector with the lowest RMSE (1.49). In dynamic respiration state dataset, the optimal configuration consists in the combination of a medium-sized bounding box, the filterless option, and the Harris corner detector with the lowest MAE (0.86) and a large bounding box, filterless option, and Harris corner detector, achieving the lowest RMSE (1.35). Overall when using the SD metric to determine the configuration options stability, Sobel filter, large bounding box and the Harris corner detector produces the lowest average values, indicating these combinations offer reliable and robust predictions across different respiration states.

## AUTHOR CONTRIBUTION

**MYAM**: Formal analysis, investigation, software, validation, writing – original draft; **MCTM**: Conceptualization, data curation, methodology, software, supervision, writing – review & editing. This author contribution format is based on Contributor Role Taxonomy (credit.niso.org).